\documentclass[10pt,twocolumn,letterpaper]{article}

\usepackage{cvpr}              % To produce the CAMERA-READY version
\usepackage{multirow}
\usepackage{enumitem}
\setlist{nosep,left=0pt}

\usepackage{xcolor}

\definecolor{mediumgreen}{RGB}{0, 128, 0} % or adjust as needed

\definecolor{cvprblue}{rgb}{0.21,0.49,0.74}
\usepackage[pagebackref,breaklinks,colorlinks,allcolors=cvprblue]{hyperref}
\usepackage{amsmath}
\usepackage{amssymb}

\def\paperID{*****} % *** Enter the Paper ID here
\def\confName{CVPR}
\def\confYear{2025}

\title{Towards Scalable SOAP Note Generation: A Weakly Supervised Multimodal Framework}

\author{
Sadia Kamal\textsuperscript{1}, Tim Oates\textsuperscript{1}, Joy Wan\textsuperscript{2} \\
\textsuperscript{1}Department of Computer Science, University of Maryland, Baltimore County \\
\textsuperscript{2}Department of Dermatology, Johns Hopkins University School of Medicine, Baltimore, MD \\
{\tt\small sadia1402@umbc.edu, oates@cs.umbc.edu, jwan7@jhmi.edu}
}

\begin{document}
\maketitle

\begin{abstract}

Skin carcinoma is the most prevalent form of cancer globally, accounting for over \$8 billion in annual healthcare expenditures. 
% Early diagnosis, accurate and timely treatment are critical to improving patient survival rates. 
In clinical settings, physicians document patient visits using detailed SOAP (Subjective, Objective, Assessment, and Plan) notes. However, manually generating these notes is labor-intensive and contributes to clinician burnout. 
% Moreover, the scarcity of annotated data, particularly in specialized domains like dermatology, limits the development of automated solutions. 
In this work, we propose a weakly supervised multimodal framework to generate clinically structured SOAP notes from limited inputs, including lesion images and sparse clinical text. 
% Captions are first generated using language models from structured features, then combined with domain-specific knowledge via retrieval-augmented generation (RAG) on a custom document to create weakly supervised SOAP notes. 
% These are used to fine-tune a Vision-LLaMA model for end-to-end structured note generation.
Our approach reduces reliance on manual annotations, enabling scalable, clinically grounded documentation while alleviating clinician burden and reducing the need for large annotated data. Our method achieves performance comparable to GPT-4o, Claude, and DeepSeek Janus Pro across key clinical relevance metrics. To evaluate clinical quality, we introduce two novel metrics MedConceptEval and Clinical Coherence Score (CCS) which assess semantic alignment with expert medical concepts and input features, respectively.
\end{abstract}

% In this work, we propose a novel weakly supervised multimodal framework that generates clinically structured SOAP notes from limited input data comprising lesion images and sparse clinical text. We first uses GPT-3.5 to generate captions from structured text features and applies retrieval-augmented generation (RAG) with a custom-curated medical knowledge base to create weakly supervised SOAP notes. These synthetic notes, paired with the corresponding image-caption inputs, are then used to fine-tune a vision-language model (LLaMA 3.2) to directly produce structured clinical notes. Our approach demonstrates that medically grounded documentation can be generated from noisy or incomplete inputs without the need for extensive manual annotations, thereby reducing clinician burden and promoting consistent, high-quality patient care through an automated, clinically relevant solution.

\vspace{-5mm}
\section{Introduction}
\label{sec:intro}

%-------------------------------------------------------------------------

% Skin cancer remains one of the most common and deadliest cancers in the United States, with approximately 9,500 new cases diagnosed daily \cite{rogers2015incidence}.
% %Among these, melanoma is the most aggressive form and continues to pose a significant public health concern. In 2024 alone, an estimated 200,340 new melanoma cases are expected, including 99,700 noninvasive and 100,640 invasive cases \cite{acs2024cancerfacts}, highlighting the urgent need for scalable and efficient diagnostic and documentation support systems.
% Clinical documentation plays a critical role in patient care, serving as the foundation for communication, diagnosis, and treatment planning.
Skin cancer remains one of the most common and deadliest cancers in the United States, with approximately 9,500 new cases diagnosed daily \cite{rogers2015incidence}, highlighting the crucial role of clinical documentation as the foundation for effective communication, accurate diagnosis, and informed treatment planning. Structured formats like SOAP (Subjective, Objective, Assessment, Plan) notes are widely adopted in the United States to ensure consistency in recording patient encounters and minimizing communication errors among healthcare professionals \cite{schloss2020towards}. However, generating these notes is labor-intensive and time-consuming, which reduces direct patient interaction time and significantly contributes to physician burnout \cite{li2024improving, biswas2024intelligent}.

Automating the generation of SOAP notes presents a promising solution to reduce administrative burden, improve documentation consistency, and allow clinicians to focus more on patient centred care. Recent advances in large language models (LLMs) have enabled impressive progress in medical natural language processing tasks, including clinical summarization, question answering \cite{singhal2023large}, lab report interpretation \cite{he2024pathclip}, and deidentification of sensitive information \cite{yang2023large}. These models can produce coherent and fluent clinical narratives, making them useful tools for medical documentation. However, general purpose LLMs often lack the domain-specific reasoning required for clinical settings, struggle to understand subtle medical context, and are generally limited to text based inputs. Their performance is further constrained in tasks such as structured note generation, especially when applied to domains like dermatology.

% Existing approaches to automated SOAP note generation, such as \cite{li2024improving}, often rely on extensive doctor-patient dialogues and large amounts of annotated training data resources that are particularly scarce in the field of dermatology and skin lesions \cite{wei2024artificial}. Moreover, developing datasets that simultaneously capture the visual characteristics of skin conditions and reflect clinically structured reasoning remains a significant challenge. To address these limitations we propose a weakly supervised, multi-modal framework for generating clinically relevant SOAP notes from limited or weakly structured data. Our method leverages structured clinical text features to generate image captions using generative pre-trained Transformer (GPT 3.5) \cite{brown2020language} and applies Retrieval-Augmented Generation (RAG) \cite{lewis2020retrieval} to extract domain-specific knowledge from a curated medical document. All available inputs, including the lesion image, GPT-generated caption, and the retrieved clinical context, are integrated to synthesize weakly supervised SOAP notes. These pseudo-labeled notes are subsequently used to fine-tune a vision-language model for the task of directly generating structured SOAP documentation from new multimodal inputs. In this study we proposed a novel method of generating structured SOAP notes with a limited data especially in the field of dermatology.

\begin{figure*}[t]
    \centering
    \includegraphics[width=0.85\textwidth]{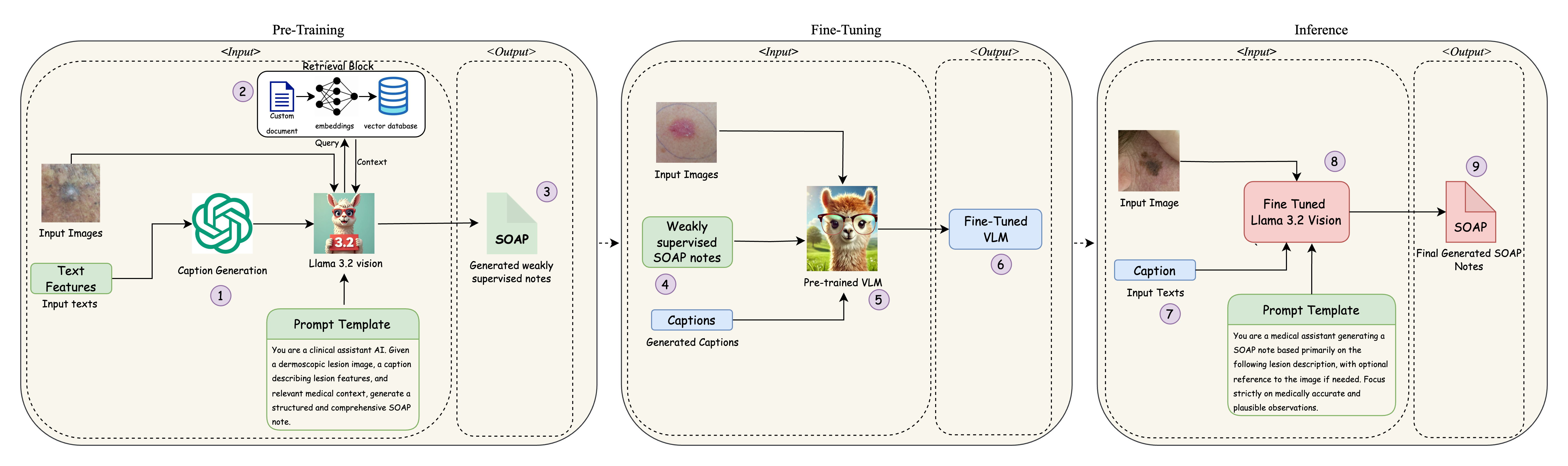}
    \caption{Overview of the proposed framework, consisting of pre-training, fine-tuning, and inference phases.}
    \label{fig:onecol}
\end{figure*}

Existing methods for automated SOAP note generation, such as K-SOAP \cite{li2024improving}, rely heavily on extensive doctor-patient dialogues and large-scale annotated datasets resources that are particularly limited in dermatology and skin lesion documentation \cite{wei2024artificial}. Moreover, capturing both the visual features of skin conditions and the underlying clinical reasoning in a structured format remains a major challenge.

To address these limitations, we propose a novel weakly supervised multimodal framework that generates structured SOAP notes from limited inputs, including lesion images and sparse clinical text. Unlike prior approaches focused solely on text-based SOAP generation or dermatologic diagnosis, our method uniquely integrates retrieval-augmented clinical knowledge, weak supervision, and multimodal synthesis to enable domain-aligned documentation without requiring large-scale annotations.

We also introduce two novel metrics to evaluate clinical and semantic quality: MedConceptEval and Clinical Coherence Score (CCS), which go beyond traditional NLP metrics by assessing alignment with defined concepts and feature consistency. By leveraging domain-guided retrieval and pseudo-labeling, our framework produces clinically relevant SOAP notes with minimal supervision, offering a scalable solution for dermatology and broader healthcare applications.
\textbf{Our main contributions are:}
\begin{itemize}
    \item We developed a weakly supervised multimodal framework for structured SOAP note generation from lesion images and limited clinical text.
    \item We introduced two novel evaluation metrics: MedConceptEval and Clinical Coherence Score (CCS) to evaluate semantic alignment with clinical concepts and input consistency.
    \item We performed statistical analysis using one-way ANOVA to quantify the effects of SOAP sections and lesion types on semantic similarity scores.
    \item We conducted qualitative evaluation using an LLM-as-a-Judge framework (Flow-Judge-v0.1) to assess structure, readability, completeness, and clinical relevance.
\end{itemize}

\vspace{-2mm}
\section{Methodology}
\label{sec:approach}

We propose a three-phase weakly supervised multimodal framework for generating clinically structured SOAP notes from limited dermatologic inputs. As illustrated in Fig.~\ref{fig:onecol}, our method consists of : (1) a pre-training phase to synthesize weakly supervised SOAP notes using generative captioning and retrieval-augmented knowledge integration, (2) a fine-tuning phase to adapt a vision-language model using the synthesized notes, and (3) an inference phase to generate high-quality structured SOAP notes from new patient data.

% \begin{figure}[h]
%     \centering
%     \includegraphics[width=0.5\textwidth]{Fig/arch-FINALfinal.drawio.png}
%     \caption{Overview of the proposed framework, consisting of pre-training, fine-tuning, and inference phases.}
%     \label{fig:onecol}
% \end{figure}

% \subsection{Dataset}
% \label{subsec:data}

% We use the PAD-UFES-20 dataset~\cite{pacheco2020pad}, which consists of 2,298 dermoscopic images along with structured metadata for 1,641 skin lesions collected from 1,373 patients. The lesions are classified into six types: \textbf{Basal Cell Carcinoma (BCC)}, \textbf{Melanoma (MEL)}, \textbf{Squamous Cell Carcinoma (SCC, including Bowen’s disease)}, and three non-cancerous conditions: \textbf{Actinic Keratosis (ACK)}, \textbf{Seborrheic Keratosis (SEK)}, and \textbf{Nevus (NEV)}.

% Approximately 58\% of the samples are confirmed through biopsy, while the remaining cases are clinically diagnosed based on expert consensus. Each lesion is linked to a CSV file containing 26 structured clinical features, which include lesion characteristics (such as size and anatomical location), patient demographics, symptom information (such as itching, bleeding, or changes in appearance), and family medical history.

\subsection{Pre-Training}
\label{subsec:pretraining}

Due to the limited availability of large-scale annotated SOAP notes in dermatology, we employed a weak supervision strategy to synthesize training data. Each sample consists a lesion image paired with structured clinical features, including lesion diameter, biopsy status, and symptom descriptors.We first utilize GPT-3.5~\cite{brown2020language} to generate a clinical caption summarizing these structured attributes into a coherent description of the lesion. To improve the clinical relevance and factual reliability of the generated notes, we design a retrieval augmented generation framework \cite{lewis2020retrieval}. The generated caption is used as a query to retrieve semantically relevant passages from a curated vector database. This database indexes document chunks extracted from authoritative medical sources, including the South Texas Skin Cancer Institute \cite{stxskincancer}, The National Cancer Institute \cite{cancer_gov}, The American Cancer Society \cite{american_cancer_society}, and The UK's National Health Service \cite{nhs_melanoma_symptoms}, covering lesion types, diagnostic criteria, symptomatology, and treatment guidelines. The retrieved context is concatenated with the original caption and provided as input to the pre-trained Vision-LLaMA 3.2 model, guided by a structured prompting template that encourages SOAP format outputs. This design addresses common limitations of pre-trained language models, such as outdated knowledge and hallucinated reasoning, and enables the model to generate reliable and clinically grounded, weakly supervised SOAP notes.

% \subsection{Fine-Tuning}
% \label{subsec:finetuning}

% We fine-tune the Vision-LLaMA 3.2 model using the synthesized dataset, where the lesion image and generated caption are treated as multimodal inputs, and the weakly supervised SOAP note serves as the training target. The model is optimized to produce structured outputs following the standard SOAP note format illustrated in Figure ~\ref{fig:SOAP_table} in Section \ref{sec:soap_structure} in the supplementary material. 

\subsection{Fine-Tuning}
\label{subsec:finetuning}

We fine-tuned the Vision-LLaMA 3.2 model using the synthesized weakly supervised dataset, where the lesion image and generated caption are treated as multimodal inputs, and the weakly supervised SOAP note serves as the training target. The model is optimized to produce structured outputs following the standard SOAP note format illustrated in Figure~\ref{fig:SOAP_table} in Section~\ref{sec:soap_structure} of the Appendix. To improve training efficiency and maintain performance, we apply two fine-tuning strategies: Parameter-Efficient Fine-Tuning (PEFT) using QLoRA, and Supervised Fine-Tuning (SFT) using weakly supervised input-output pairs. Detailed training setup and method variants (PEFT and SFT) are provided in the Appendix ~\ref{sec:appendix method}.

% Detailed training setup and dataset information can be found in the Appendix, specifically in Section.

% \subsection{Training Setup}
% \label{subsec:training_setup}

% The fine-tuning of the Vision-LLaMA 3.2 model is performed using Quantized Low-Rank Adaptation (QLoRA) with a low-rank dimension \( r = 8 \), a scaling factor \( \alpha = 16 \), and no dropout applied. LoRA modules are inserted into the query, key, value, output, gate, up, and down projections within the model's transformer blocks. Supervised Fine-Tuning (SFT) is conducted with a batch size of 8, using gradient accumulation over 4 steps and 10 warmup steps at the beginning of training. Fine-tuning is performed for 500 epochs with a linear learning rate scheduler, starting from an initial learning rate of \( 2 \times 10^{-4} \). The optimizer used is AdamW with 8-bit precision to improve memory efficiency. Pretraining to generate weakly supervised SOAP notes took approximately 9 hours, while the final fine-tuning process was completed in 1.5 hours on an NVIDIA A100 GPU with 80 GB of VRAM. Mixed-precision training with bfloat16 (bf16) format was employed to optimize memory utilization throughout the training process.

\subsection{Inference}
\label{subsec:inference}

At inference time, the fine-tuned Vision-LLaMA model receives a lesion image along with its corresponding clinical features, which were first converted into a clinical caption. The model then generates a structured SOAP note. Because the model has been fine-tuned on weakly supervised yet clinically reliable data, it generalizes effectively to new cases. This enables scalable and structured generation of dermatology SOAP notes, even in environments where expert annotations are limited or unavailable.

\vspace{-2mm}
\section{Evaluation}

We evaluated the generated SOAP notes using both quantitative and qualitative methods. For quantitative evaluation we compare against annotated ground truth using standard and clinical domain NLP metrics. Additionally, we introduced two novel clinical relevance metrics: \textbf{MedConceptEval} and \textbf{Clinical Coherence Score (CCS)}. 
% They evaluate the semantic alignment between SOAP notes and disease-specific concept sets, and caption using Clinical BERT, respectively dimensions that not captured by traditional NLP metrics. 
The qualitative evaluation is based on the LLM-as-a-Judge approach \cite{zheng2023judging} with Flow-Judge-v0.1, where the judge model evaluates structure, readability, medical relevance, and completeness with conventional SOAP note standards.

\subsection{Quantitative Evaluation}

% \subsection{Structure}
% We verify that generated outputs adhere to the SOAP (Subjective, Objective, Assessment, Plan) format. This ensures that each clinical note is correctly segmented, enabling interpretability, retrievability, and consistency with Electronic Health Record (EHR) documentation standards.
% \subsection{Structure}
% We verify that the generated outputs follow the SOAP (Subjective, Objective, Assessment, Plan) format. This ensures that each clinical note is correctly segmented, enabling interpretability, retrievability, and consistency with Electronic Health Record (EHR) documentation standards.

% \subsection{Readability}
% We assess readability using standard metrics such as the Flesch-Kincaid Grade Level, Gunning Fog Index, and SMOG Index to determine if the generated text is appropriate for clinical interpretation. This ensures that outputs are understandable to practitioners while maintaining clinical detail.

\subsubsection{MedConceptEval}
% To measure clinical relevance, we propose MedConceptEval, a semantic evaluation framework that assesses the alignment of each SOAP note section with clinically authoritative concept sets curated for six different dermatological classes from . We use SciSpaCy to extract these medical concepts and create disease-specific keyword sets, called descriptor banks. Each SOAP section is then encoded using Clinical BERT and cosine similarity is computed against the relevant concept sets. We calculate both the average and maximum similarity scores for each section of the SOAP across five cases per dermatological class, identifying the best-matching clinical concepts. This automated pipeline enables robust, section-wise evaluation of SOAP notes at a semantic level, ensuring that generated notes accurately capture clinically relevant features beyond simple keyword matching.

To evaluate clinical relevance, we introduce MedConceptEval, a semantic evaluation framework designed to assess the alignment of each SOAP note section with clinically validated concept sets. These concept sets, referred to as descriptor banks, are curated for six major dermatological classes derived from reputable clinical resources \cite{mayoclinic_melanoma_symptoms}. A language model is employed to extract relevant medical concepts and construct disease specific keyword sets. Each section of the generated SOAP note is encoded using ClinicalBERT, and cosine similarity is computed against the corresponding descriptor bank. For each dermatological class, we calculate both average and maximum similarity scores per section across five representative cases. This method provides a robust, interpretable, and clinically grounded evaluation of the generated SOAP notes, ensuring alignment with disease specific clinical terminology beyond surface level keyword matching.

% \begin{table}[htbp]
% \renewcommand{\arraystretch}{1.1}
% \centering
% \footnotesize
% \begin{tabular}{|p{1.3 cm}|l|c|c|}
% \hline
% \textbf{Condition} & \textbf{Section} & \textbf{Avg Similarity} & \textbf{Max Similarity} \\
% \hline
% \multirow{4}{=}{Seborrheic Keratosis (SEK)} 
% & Subjective & 0.7768 & 0.8746 \\
% & Objective  & 0.7952 & 0.8648 \\
% & Assessment & 0.8168 & 0.8680 \\
% & Plan       & 0.7764 & 0.8310 \\
% \hline

% \multirow{4}{=}{Nevus (NEV)} 
% & Subjective & 0.7786 & 0.8468 \\
% & Objective  & 0.7786 & 0.8598 \\
% & Assessment & 0.8006 & 0.8708 \\
% & Plan       & 0.8626 & 0.8976 \\
% \hline

% \multirow{4}{=}{Melanoma (MEL)} 
% & Subjective & 0.7790 & 0.8624 \\
% & Objective  & 0.7952 & 0.8676 \\
% & Assessment & 0.8234 & 0.8770 \\
% & Plan       & 0.8526 & 0.9036 \\
% \hline

% \multirow{4}{=}{Actinic Keratosis (ACK)} 
% & Subjective & 0.7354 & 0.8182 \\
% & Objective  & 0.7844 & 0.8554 \\
% & Assessment & 0.8092 & 0.8458 \\
% & Plan       & 0.8400 & 0.8864 \\
% \hline

% \multirow{4}{=}{Squamous Cell Carcinoma (SCC)} 
% & Subjective & 0.7846 & 0.8360 \\
% & Objective  & 0.7754 & 0.8360 \\
% & Assessment & 0.7802 & 0.8398 \\
% & Plan       & 0.7854 & 0.8596 \\
% \hline

% \multirow{4}{=}{Basal Cell Carcinoma (BCC)} 
% & Subjective & 0.7740 & 0.8464 \\
% & Objective  & 0.7658 & 0.8182 \\
% & Assessment & 0.7882 & 0.8220 \\
% & Plan       & 0.7738 & 0.8212 \\
% \hline
% \end{tabular}
% \caption{\textbf{MedConceptEval:} Semantic similarity between SOAP sections and curated clinical concept sets across six dermatologic conditions.}
% \label{tab:medconcepteval}
% \end{table}

\begin{table}[htbp]
\centering
\scriptsize  % You can change this to \tiny for even smaller
\renewcommand{\arraystretch}{1.1}
\resizebox{0.85\linewidth}{!}{  % Resize table to 95% of text width
\begin{tabular}{|p{1.5cm}|l|c|c|}
\hline
\textbf{Condition} & \textbf{Section} & \textbf{Avg Similarity} & \textbf{Max Similarity} \\
\hline
\multirow{4}{=}{Seborrheic Keratosis (SEK)} 
& Subjective & 0.7768 & 0.8746 \\
& Objective  & 0.7952 & 0.8648 \\
& Assessment & 0.8168 & 0.8680 \\
& Plan       & 0.7764 & 0.8310 \\
\hline

\multirow{4}{=}{Nevus (NEV)} 
& Subjective & 0.7786 & 0.8468 \\
& Objective  & 0.7786 & 0.8598 \\
& Assessment & 0.8006 & 0.8708 \\
& Plan       & 0.8626 & 0.8976 \\
\hline

\multirow{4}{=}{Melanoma (MEL)} 
& Subjective & 0.7790 & 0.8624 \\
& Objective  & 0.7952 & 0.8676 \\
& Assessment & 0.8234 & 0.8770 \\
& Plan       & 0.8526 & 0.9036 \\
\hline

\multirow{4}{=}{Actinic Keratosis (ACK)} 
& Subjective & 0.7354 & 0.8182 \\
& Objective  & 0.7844 & 0.8554 \\
& Assessment & 0.8092 & 0.8458 \\
& Plan       & 0.8400 & 0.8864 \\
\hline

\multirow{4}{=}{Squamous Cell Carcinoma (SCC)} 
& Subjective & 0.7846 & 0.8360 \\
& Objective  & 0.7754 & 0.8360 \\
& Assessment & 0.7802 & 0.8398 \\
& Plan       & 0.7854 & 0.8596 \\
\hline

\multirow{4}{=}{Basal Cell Carcinoma (BCC)} 
& Subjective & 0.7740 & 0.8464 \\
& Objective  & 0.7658 & 0.8182 \\
& Assessment & 0.7882 & 0.8220 \\
& Plan       & 0.7738 & 0.8212 \\
\hline
\end{tabular}
}
\caption{\textbf{MedConceptEval:} Semantic similarity between SOAP sections and curated clinical concept sets across six dermatologic conditions.}
\label{tab:medconcepteval}
\end{table}

\begin{table*}[t]
\centering
\small
\resizebox{0.85\textwidth}{!}{%
\begin{tabular}{|l|cccc|cccc|cccc|}
\hline
\textbf{Metrics} & \multicolumn{4}{c|}{\textbf{Case 1}} & \multicolumn{4}{c|}{\textbf{Case 2}} & \multicolumn{4}{c|}{\textbf{Case 3}} \\
\cline{2-13}
& Ours & GPT-4o & Janus Pro & Claude & Ours & GPT-4o & Janus Pro & Claude & Ours & GPT-4o & Janus Pro & Claude \\
\hline
ROUGE-1 \cite{lin2004rouge}      & \textcolor{mediumgreen}{0.396} & 0.324 & 0.348 & 0.294 & 0.4183 & 0.399 & 0.397 & 0.460 & 0.3999 & 0.4360 & 0.351 & 0.480 \\
ROUGE-2 \cite{lin2004rouge}       & 0.0827 & 0.0326 & 0.1228 & 0.0281 & \textcolor{mediumgreen}{0.125} & 0.0966 & 0.1122 & 0.0671 & 0.0939 & 0.0909 & 0.1088 & 0.0805 \\
ROUGE-L \cite{lin2004rouge}       & 0.1748 & 0.1943 & 0.2347 & 0.1538 & \textcolor{mediumgreen}{0.2614} & 0.2214 & 0.2299 & 0.2266 & 0.2181 & 0.2556 & 0.2229 & 0.2472 \\
% BLEU            & 0.0188 & 0.0053 & 0.0283 & 0.0094 & \textcolor{blue}{0.0583} & 0.0402 & 0.0449 & 0.0495 & 0.0195 & 0.0278 & 0.0386 & 0.0290 \\
METEOR \cite{lewis2020retrieval}         & \textcolor{mediumgreen}{0.2221} & 0.1745 & 0.1804 & 0.1952 & \textcolor{mediumgreen}{0.2495} & 0.1692 & 0.1728 & 0.2202 & 0.2276 & 0.2242 & 0.1968 & 0.2370 \\
CHRF++ \cite{popovic2017chrf++}         & \textcolor{mediumgreen}{44.91} & 44.90 & 37.25 & 38.92 & \textcolor{mediumgreen}{43.78} & 42.39 & 41.64 & 42.30 & \textcolor{mediumgreen}{47.515} & 45.71 & 41.75 & 47.33 \\
BERT (F1) \cite{zhang2019bertscore} & \textcolor{mediumgreen}{0.1223} & 0.0619 & 0.1144 & 0.0117 & 0.0974 & 0.1487 & 0.1460 & 0.1481 & 0.0770 & 0.2100 & 0.0994 & 0.2105 \\
ClinicalBERT (F1) \cite{huang2019clinicalbert} & \textcolor{mediumgreen}{0.7750} & 0.7409 & 0.7348 & 0.6974 & 0.7609 & 0.7723 & 0.7550 & 0.7528 & 0.7890 & 0.8098 & 0.7573 & 0.7846 \\
\hline
\end{tabular}
}
\caption{SOAP note evaluation across three cases using lexical, character-level and semantic metrics.}

% \caption{Evaluation of generated SOAP notes across lexical (ROUGE, METEOR), character-level (CHRF++), and semantic F1-score (BERT, Clinical BERT) metrics for three cases. Models: \textbf{Our Approach (Ours)}, GPT-4o, Janus Pro Deepseek, Claude 3.7 Sonnet.}
\label{tab:all_model_eval_combined}
\end{table*}

Table \ref{tab:medconcepteval} shows the MedConceptEval results. Overall, the Assessment and Plan sections consistently achieved higher average similarity scores compared to Subjective and Objective sections, indicating stronger alignment with medically relevant concepts. Conditions like Melanoma and Nevus demonstrated particularly high alignment, with maximum similarity values exceeding 0.90 in the Plan section, suggesting that generated notes closely matched expert medical descriptors. However, slightly lower scores were observed for SCC and BCC, highlighting potential areas for improvement in capturing subtle clinical features for these cases.
% \subsubsection{Statistical Significance Analysis}

% We conducted two separate one-way ANOVA tests to examine the effects of SOAP note sections and skin lesion types on the average similarity scores. The results showed that the SOAP section had a statistically significant effect on similarity (F(3, 20) = 3.88, p = 0.024), indicating that the alignment between the generated notes and medically relevant concepts varied across sections such as Subjective, Objective, Assessment, and Plan. In contrast, the skin lesion classes did not have a significant effect (F(5, 18) = 0.93, p = 0.487), suggesting that the model maintained relatively consistent performance across different disease categories, including SEK, Nevus, Melanoma, ACK, SCC, and BCC. These findings highlight that the structure of the SOAP note plays a more critical role in influencing semantic similarity than the specific lesion type.

\textbf{Statistical Significance Analysis}: We performed two separate one-way ANOVA tests to assess the impact of SOAP note sections and dermatological conditions on the average similarity scores. The results showed a statistically significant effect of the SOAP section \((F(3, 20) = 3.88, p = 0.024)\), indicating that semantic alignment varied across different sections (Subjective, Objective, Assessment, and Plan). In contrast, the lesion type had no significant effect \((F(5, 18) = 0.93, p = 0.487)\), suggesting that the model maintained consistent performance across different dermatological conditions. Here, \(F(d_1, d_2)\) denotes the F-statistic with \(d_1\) degrees of freedom between groups (number of categories minus one) and \(d_2\) degrees of freedom within groups (total observations minus the number of categories). These findings suggest that the organization of information within the SOAP structure plays a greater role in influencing semantic quality than the specific disease category.

\subsubsection{Clinical Coherence Score (CCS)}
We introduce the Clinical Coherence Score (CCS), a metric that evaluates the semantic alignment between the caption derived from input features and the structured SOAP note sections. To compute this score, we use ClinicalBERT based contextual embeddings, which are specifically designed to capture clinical terminology and relationships. Both the caption and each SOAP section are embedded into a shared clinical semantic space, and the cosine similarity between these embeddings is calculated to assess how well the generated note aligns with the corresponding caption. Table \ref{tab:semantic_alignment_scores} shows strong semantic alignment across all cases, with Case 3 scoring highest. The Subjective and Assessment sections show the strongest alignment, while the Plan section shows slightly lower coherence, likely because plan specific keywords are not present in the caption and are instead entirely generated by the LLM.

% \begin{table}[h]
% \centering
% \small
% \begin{tabular}{|l|c|c|c|c|c|}
% \hline
% \textbf{SOAP Sections} & \textbf{Case 1} & \textbf{Case 2} & \textbf{Case 3} \\
% \hline
% Subjective & 0.9308 & 0.9306 & 0.9318 \\
% \hline
% Objective  & 0.9168 & 0.8891 & 0.9334 \\
% \hline
% Assessment & 0.9178 & 0.8933 & 0.9208 \\
% \hline
% Plan       & 0.8795 & 0.8842 & 0.8981 \\
% \hline
% \textbf{Average Score} & \textbf{0.9112} & \textbf{0.8993} & \textbf{0.9210} \\
% \hline
% \end{tabular}
% \caption{\textbf{Clinical Coherence Score:} Semantic alignment between caption and SOAP note section across three lesion images.}
% \label{tab:semantic_alignment_scores}
% \end{table}

\begin{table}[h]

\centering
\scriptsize  % You can also use \tiny for further size reduction

\setlength{\tabcolsep}{3pt} % default is 6pt

\renewcommand{\arraystretch}{1.1}
\resizebox{0.7\linewidth}{!}{  % Adjust the width percentage as needed
\begin{tabular}{|l|c|c|c|}
\hline
\textbf{Sections} & \textbf{Case 1} & \textbf{Case 2} & \textbf{Case 3} \\
\hline
Subjective & 0.9308 & 0.9306 & 0.9318 \\
\hline
Objective  & 0.9168 & 0.8891 & 0.9334 \\
\hline
Assessment & 0.9178 & 0.8933 & 0.9208 \\
\hline
Plan       & 0.8795 & 0.8842 & 0.8981 \\
\hline
\textbf{Average Score} & \textbf{0.9112} & \textbf{0.8993} & \textbf{0.9210} \\
\hline
\end{tabular}
}
\caption{\textbf{Clinical Coherence Score:} Semantic alignment between caption and SOAP note sections across three lesion images.}
\label{tab:semantic_alignment_scores}
\end{table}

Furthermore, we obtained expert annotated SOAP notes for three different lesion images from a board-certified dermatologist. Each lesion image, along with its corresponding caption, was provided to the dermatologist to generate structured SOAP notes. These expert written notes were treated as ground truth (reference) and are compared with our generated notes (candidate) to evaluate their alignment using a range of lexical, character and semantic metrics, including ROUGE, METEOR, CHRF++, BERT and ClinicalBERT Scores. For more details please refer to Appendix ~\ref{sec:appendix quat}

In Table~\ref{tab:all_model_eval_combined}, our model demonstrates consistently strong performance, especially on METEOR and CHRF++, indicating high fluency and surface-level coherence. While GPT-4o achieved slightly higher ROUGE-L in Cases 1 and 3, our method outperformed it in ROUGE-2 and METEOR, which better reflect phrasal and semantic overlap. Notably, our ClinicalBERT F1 scores are either the highest or on par with the best-performing model in each case, underscoring our model's superior alignment with clinical concepts. Here, the difference between the BERT and ClinicalBERT scores further highlights how clinically tuned models capture semantic relevance more effectively than general-purpose transformer models. Overall, our results are comparable with the other models and these results confirm the effectiveness of our weakly supervised multimodal approach for producing high-quality and clinically grounded SOAP notes.

% In the Table \ref{tab:all_model_eval_combined}, our approach consistently outperforms GPT-4o, Janus Pro, and Claude across most evaluation dimensions. In particular, our method achieved a Clinical BERT F1 score of 0.7750, 0.7609, and 0.7890 across Case 1, Case 2, and Case 3, respectively, demonstrating superior and comparable alignment with clinical concepts compared to baseline models. Similarly, we observe strong performance on CHRF++ and METEOR, indicating that our generated notes maintain both character-level and sentence-level coherence. Although GPT-4o showed slightly higher ROUGE-L scores on Case 1 and Case 3, our method demonstrated better overall completeness and medical fidelity, as reflected by higher ClinicalBERT and BERTScore values. This highlights that our framework not only generates linguistically coherent notes but also preserves medically relevant information, which is critical for clinical documentation tasks. The difference between BERT and ClinicalBERT scores shows how the semantic aligment varies when evaluating with the model trained on clinical data comparing to general transformer model.  Overall, our results are comparable with the other models and these results confirm the effectiveness of our weakly supervised multimodal approach for producing high-quality and clinically grounded SOAP notes.

\subsection{Qualitative Evaluation}
\subsubsection{LLM-as-a-Judge}

% We conducted a qualitative evaluation using a \textbf{LLM-as-Judge} framework, since getting a human verifier is expensive and there is and time consuming to scoring all the notes. Therefore, in this study we employed \textbf{Flow-Judge-v0.1} to assess each generated SOAP note across four criteria. In the HuggingFace Judge Arena: Benchmarking LLMs as Evaluators, Flow-Judge by Flow AI is the best judge model with the ELO rating of 1335, competing larger models such as GPT-4o, Meta Llama 3.1, and Claude 3 Opus with the ELO ratings of 1312, 1289, and 1268. In our study, each generated note was rated by Flow-Judge on a 5-point Likert scale (5 = Excellent, 1 = Poor) based on the following evaluation criteria:

We conducted a qualitative evaluation using an LLM-as-Judge framework, as obtaining human verifiers is both time-consuming and resource-intensive. Specifically, we employed \textbf{Flow-Judge-v0.1} to assess each generated SOAP note across four criteria. According to the HuggingFace Judge Arena: Benchmarking LLMs as Evaluators, Flow-Judge (3.8B) is an open-source model developed by Flow AI, achieves the highest ELO rating of 1335, outperforming larger proprietary models such as GPT-4o (1320), Claude 3 Opus (1268), and Meta Llama 3.1 405B (1267). In this study, Flow-Judge rated each note on a 5-point rating scale (1 = Poor, 5 = Excellent) based on the following evaluation criteria: Structure, Readability, Completeness, Medical Relevance. 

%\begin{itemize} \item \textbf{Structure:} Does the clinical note correctly follow the structured SOAP format, with distinct and appropriate content under each section (Subjective: Chief Complaint and Medical History; Objective: Examination findings and Observations; Assessment: Investigations, Diagnosis, and Summary; Plan: Treatment Plan and Patient Education)? \item \textbf{Readability:} Is the language of the clinical note clear, concise, and readable for a medical professional without excessive complexity or ambiguity? \item \textbf{Completeness:} Does the clinical note cover all the key details described in the input lesion description and context, and address all aspects of the clinical scenario, ensuring that no critical details are overlooked? \item \textbf{Medical Relevance:} Is the clinical content of the SOAP note medically relevant, plausible, and appropriate given the input description of the skin lesion? \end{itemize}

% \begin{figure*}[t]
%     \centering
%     \includegraphics[width=\textwidth]{Fig/flowjudge.drawio (1).png}
%     \caption{Comparison of Flow-Judge Feedback Across Four Language Models}
%     \label{fig:LLM}
% \end{figure*}

This setup enabled a structured, blinded review of the generated outputs, allowing us to capture both technical and clinical quality dimensions beyond traditional quantitative scores. The total scores for each model were shown in Fig \ref{fig:LLM} ~\ref{sec:appendix qual} of the Appendix, our approach achieved a perfect score of 20/20, with feedback highlighting its clear organization, clinical accuracy, and comprehensive coverage of all relevant details. In contrast, GPT-4o and the other baselines scored between 18/20 and 19/20. While their outputs were well-structured and readable, minor issues such as omissions of specific symptoms or less detailed assessments slightly impacted their completeness scores. Overall, Flow-Judge evaluations confirm that our framework delivers higher clinical fidelity and completeness than SOTA models, highlighting key strengths and improvement areas. 
% Please refer to Appendix ~\ref{sec:Flow judge} for detailed criteria. 
% For detailed model comparisons, including corresponding feedback and ratings, please refer to Appendix ~\ref{sec:appendix qual}.

% You must include your signed IEEE copyright release form when you submit your finished paper.
% We MUST have this form before your paper can be published in the proceedings.

% Please direct any questions to the production editor in charge of these proceedings at the IEEE Computer Society Press:
% \url{https://www.computer.org/about/contact}.
% \input{5_limit}

\section{Conclusion}
\vspace{-2mm}
\label{sec:conc}

In this work, we presented a weakly supervised multimodal framework for generating clinically structured SOAP notes from limited dermatologic inputs. By leveraging generative language model generated captions, retrieval-augmented knowledge integration, and fine-tuning a Vision-LLaMA model with weak supervision, this study reduces dependence on large-scale expert annotations while maintaining strong clinical relevance and structural coherence. Through both qualitative and quantitative evaluations, including our proposed metrics such as MedConceptEval and Clinical Coherence Score (CCS), we demonstrated that our method produces high-quality, clinically meaningful notes, advancing the development of scalable and reliable clinical documentation systems. Ultimately, our framework has the potential to accelerate dermatology clinical workflows, reduce time-to treatment, and improve overall patient care. For limitations and future work please refer to Appendix Section ~\ref{sec:limit}.

\label{sec:conc}
{
    \small
    \bibliographystyle{ieeenat_fullname}
    \bibliography{main}

\begin{thebibliography}{32}
\providecommand{\natexlab}[1]{#1}
\providecommand{\url}[1]{\texttt{#1}}
\expandafter\ifx\csname urlstyle\endcsname\relax
  \providecommand{\doi}[1]{doi: #1}\else
  \providecommand{\doi}{doi: \begingroup \urlstyle{rm}\Url}\fi

\bibitem[Amenyo et~al.(2025)Amenyo, Grossman, Brown, and Wylie-Toal]{amenyo2025assessment}
Solomon Amenyo, Maura~R Grossman, Daniel~G Brown, and Brendan Wylie-Toal.
\newblock Assessment of ai-generated pediatric rehabilitation soap-note quality.
\newblock \emph{arXiv preprint arXiv:2503.15526}, 2025.

\bibitem[{American Cancer Society}(2024)]{american_cancer_society}
{American Cancer Society}.
\newblock Donate to support the fight against cancer.
\newblock \url{https://donate.cancer.org/}, 2024.

\bibitem[Biswas and Talukdar(2024)]{biswas2024intelligent}
Anjanava Biswas and Wrick Talukdar.
\newblock Intelligent clinical documentation: Harnessing generative ai for patient-centric clinical note generation.
\newblock \emph{arXiv preprint arXiv:2405.18346}, 2024.

\bibitem[Brown et~al.(2020)Brown, Mann, Ryder, Subbiah, Kaplan, Dhariwal, Neelakantan, Shyam, Sastry, Askell, et~al.]{brown2020language}
Tom Brown, Benjamin Mann, Nick Ryder, Melanie Subbiah, Jared~D Kaplan, Prafulla Dhariwal, Arvind Neelakantan, Pranav Shyam, Girish Sastry, Amanda Askell, et~al.
\newblock Language models are few-shot learners.
\newblock \emph{Advances in neural information processing systems}, 33:\penalty0 1877--1901, 2020.

\bibitem[Chen and Hirschberg(2024)]{chen2024exploring}
Yu-Wen Chen and Julia Hirschberg.
\newblock Exploring robustness in doctor-patient conversation summarization: An analysis of out-of-domain soap notes.
\newblock \emph{arXiv preprint arXiv:2406.02826}, 2024.

\bibitem[Cheng et~al.(2025)Cheng, Song, Fan, Ma, Sun, Xu, Yan, Chen, Zhang, and Chen]{cheng2025caparena}
Kanzhi Cheng, Wenpo Song, Jiaxin Fan, Zheng Ma, Qiushi Sun, Fangzhi Xu, Chenyang Yan, Nuo Chen, Jianbing Zhang, and Jiajun Chen.
\newblock Caparena: Benchmarking and analyzing detailed image captioning in the llm era.
\newblock \emph{arXiv preprint arXiv:2503.12329}, 2025.

\bibitem[Dettmers et~al.(2023)Dettmers, Pagnoni, Holtzman, and Zettlemoyer]{dettmers2023qlora}
Tim Dettmers, Artidoro Pagnoni, Ari Holtzman, and Luke Zettlemoyer.
\newblock Qlora: Efficient finetuning of quantized llms.
\newblock \emph{Advances in neural information processing systems}, 36:\penalty0 10088--10115, 2023.

\bibitem[He et~al.(2024)He, Liu, Yang, Chen, Hammer, Xu, and Popescu]{he2024pathclip}
Fei He, Kai Liu, Zhiyuan Yang, Yibo Chen, Richard~D Hammer, Dong Xu, and Mihail Popescu.
\newblock pathclip: Detection of genes and gene relations from biological pathway figures through image-text contrastive learning.
\newblock \emph{IEEE Journal of Biomedical and Health Informatics}, 2024.

\bibitem[Huang et~al.(2019)Huang, Altosaar, and Ranganath]{huang2019clinicalbert}
Kexin Huang, Jaan Altosaar, and Rajesh Ranganath.
\newblock Clinicalbert: Modeling clinical notes and predicting hospital readmission.
\newblock \emph{arXiv preprint arXiv:1904.05342}, 2019.

\bibitem[Jung et~al.(2024)Jung, Kim, Choi, Seo, Kim, Han, Kee, Park, Ko, Kim, et~al.]{jung2024enhancing}
HyoJe Jung, Yunha Kim, Heejung Choi, Hyeram Seo, Minkyoung Kim, JiYe Han, Gaeun Kee, Seohyun Park, Soyoung Ko, Byeolhee Kim, et~al.
\newblock Enhancing clinical efficiency through llm: Discharge note generation for cardiac patients.
\newblock \emph{arXiv preprint arXiv:2404.05144}, 2024.

\bibitem[Leong et~al.(2024)Leong, Gao, Shuai, Zhang, and Pamuksuz]{leong2024efficient}
Hui~Yi Leong, Yi~Fan Gao, Ji Shuai, Yang Zhang, and Uktu Pamuksuz.
\newblock Efficient fine-tuning of large language models for automated medical documentation.
\newblock \emph{arXiv preprint arXiv:2409.09324}, 2024.

\bibitem[Lewis et~al.(2020)Lewis, Perez, Piktus, Petroni, Karpukhin, Goyal, K{\"u}ttler, Lewis, Yih, Rockt{\"a}schel, et~al.]{lewis2020retrieval}
Patrick Lewis, Ethan Perez, Aleksandra Piktus, Fabio Petroni, Vladimir Karpukhin, Naman Goyal, Heinrich K{\"u}ttler, Mike Lewis, Wen-tau Yih, Tim Rockt{\"a}schel, et~al.
\newblock Retrieval-augmented generation for knowledge-intensive nlp tasks.
\newblock \emph{Advances in neural information processing systems}, 33:\penalty0 9459--9474, 2020.

\bibitem[Li et~al.(2024)Li, Wu, Smith, Lo, and Liu]{li2024improving}
Yizhan Li, Sifan Wu, Christopher Smith, Thomas Lo, and Bang Liu.
\newblock Improving clinical note generation from complex doctor-patient conversation.
\newblock \emph{arXiv preprint arXiv:2408.14568}, 2024.

\bibitem[Lin(2004)]{lin2004rouge}
Chin-Yew Lin.
\newblock Rouge: A package for automatic evaluation of summaries.
\newblock In \emph{Text summarization branches out}, pages 74--81, 2004.

\bibitem[{Mayo Clinic Staff}(2023)]{mayoclinic_melanoma_symptoms}
{Mayo Clinic Staff}.
\newblock Melanoma - symptoms and causes, 2023.
\newblock Accessed: 2025-04-30.

\bibitem[{National Cancer Institute}(2024)]{cancer_gov}
{National Cancer Institute}.
\newblock Comprehensive cancer information.
\newblock \url{https://www.cancer.gov/}, 2024.

\bibitem[{NHS}(2024)]{nhs_melanoma_symptoms}
{NHS}.
\newblock Melanoma skin cancer - symptoms.
\newblock \url{https://www.nhs.uk/conditions/melanoma-skin-cancer/symptoms/}, 2024.

\bibitem[Pacheco et~al.(2020)Pacheco, Lima, Salomao, Krohling, Biral, de~Angelo, Alves~Jr, Esgario, Simora, Castro, et~al.]{pacheco2020pad}
Andre~GC Pacheco, Gustavo~R Lima, Amanda~S Salomao, Breno Krohling, Igor~P Biral, Gabriel~G de Angelo, F{\'a}bio~CR Alves~Jr, Jos{\'e}~GM Esgario, Alana~C Simora, Pedro~BC Castro, et~al.
\newblock Pad-ufes-20: A skin lesion dataset composed of patient data and clinical images collected from smartphones.
\newblock \emph{Data in brief}, 32:\penalty0 106221, 2020.

\bibitem[Popovi{\'c}(2017)]{popovic2017chrf++}
Maja Popovi{\'c}.
\newblock chrf++: words helping character n-grams.
\newblock In \emph{Proceedings of the second conference on machine translation}, pages 612--618, 2017.

\bibitem[Ramprasad et~al.(2023)Ramprasad, Ferracane, and Selvaraj]{ramprasad2023generating}
Sanjana Ramprasad, Elisa Ferracane, and Sai~P Selvaraj.
\newblock Generating more faithful and consistent soap notes using attribute-specific parameters.
\newblock In \emph{Machine Learning for Healthcare Conference}, pages 631--649. PMLR, 2023.

\bibitem[Rogers et~al.(2015)Rogers, Weinstock, Feldman, and Coldiron]{rogers2015incidence}
Howard~W. Rogers, Martin~A. Weinstock, Steven~R. Feldman, and Brett~M. Coldiron.
\newblock Incidence estimate of nonmelanoma skin cancer (keratinocyte carcinomas) in the us population, 2012.
\newblock \emph{JAMA Dermatology}, 151\penalty0 (10):\penalty0 1081--1086, 2015.

\bibitem[Schloss and Konam(2020)]{schloss2020towards}
Benjamin Schloss and Sandeep Konam.
\newblock Towards an automated soap note: classifying utterances from medical conversations.
\newblock In \emph{Machine Learning for Healthcare Conference}, pages 610--631. PMLR, 2020.

\bibitem[Singh et~al.(2023)Singh, Pan, Andres-Ferrer, Del-Agua, Diehl, Pinto, and Vozila]{singh2023large}
Gagandeep Singh, Yue Pan, Jesus Andres-Ferrer, Miguel Del-Agua, Frank Diehl, Joel Pinto, and Paul Vozila.
\newblock Large scale sequence-to-sequence models for clinical note generation from patient-doctor conversations.
\newblock In \emph{Proceedings of the 5th Clinical Natural Language Processing Workshop}, pages 138--143, 2023.

\bibitem[Singhal et~al.(2023)Singhal, Azizi, Tu, Mahdavi, Wei, Chung, Scales, Tanwani, Cole-Lewis, Pfohl, et~al.]{singhal2023large}
Karan Singhal, Shekoofeh Azizi, Tao Tu, S~Sara Mahdavi, Jason Wei, Hyung~Won Chung, Nathan Scales, Ajay Tanwani, Heather Cole-Lewis, Stephen Pfohl, et~al.
\newblock Large language models encode clinical knowledge.
\newblock \emph{Nature}, 620\penalty0 (7972):\penalty0 172--180, 2023.

\bibitem[{Skin Cancer Specialists}(2024)]{stxskincancer}
{Skin Cancer Specialists}.
\newblock Skin cancer diagnosis and treatment.
\newblock \url{https://www.stxskincancer.com}, 2024.

\bibitem[Van~Veen et~al.(2023)Van~Veen, Van~Uden, Blankemeier, Delbrouck, Aali, Bluethgen, Pareek, Polacin, Reis, Seehofnerova, et~al.]{van2023clinical}
Dave Van~Veen, Cara Van~Uden, Louis Blankemeier, Jean-Benoit Delbrouck, Asad Aali, Christian Bluethgen, Anuj Pareek, Malgorzata Polacin, Eduardo~Pontes Reis, Anna Seehofnerova, et~al.
\newblock Clinical text summarization: adapting large language models can outperform human experts.
\newblock \emph{Research square}, pages rs--3, 2023.

\bibitem[Wei et~al.(2024)Wei, Tada, So, and Torres]{wei2024artificial}
Maria~L Wei, Mikio Tada, Alexandra So, and Rodrigo Torres.
\newblock Artificial intelligence and skin cancer.
\newblock \emph{Frontiers in medicine}, 11:\penalty0 1331895, 2024.

\bibitem[Yan et~al.(2024)Yan, Niu, Li, Zhang, Yin, Fei, Peng, Bi, Feng, Chen, et~al.]{yan2024large}
Lawrence~KQ Yan, Qian Niu, Ming Li, Yichao Zhang, Caitlyn~Heqi Yin, Cheng Fei, Benji Peng, Ziqian Bi, Pohsun Feng, Keyu Chen, et~al.
\newblock Large language model benchmarks in medical tasks.
\newblock \emph{arXiv preprint arXiv:2410.21348}, 2024.

\bibitem[Yang et~al.(2023)Yang, Tan, Lu, Thirunavukarasu, Ting, and Liu]{yang2023large}
Rui Yang, Ting~Fang Tan, Wei Lu, Arun~James Thirunavukarasu, Daniel Shu~Wei Ting, and Nan Liu.
\newblock Large language models in health care: Development, applications, and challenges.
\newblock \emph{Health Care Science}, 2\penalty0 (4):\penalty0 255--263, 2023.

\bibitem[Zhang et~al.(2019)Zhang, Kishore, Wu, Weinberger, and Artzi]{zhang2019bertscore}
Tianyi Zhang, Varsha Kishore, Felix Wu, Kilian~Q Weinberger, and Yoav Artzi.
\newblock Bertscore: Evaluating text generation with bert.
\newblock \emph{arXiv preprint arXiv:1904.09675}, 2019.

\bibitem[Zheng et~al.(2023)Zheng, Chiang, Sheng, Zhuang, Wu, Zhuang, Lin, Li, Li, Xing, et~al.]{zheng2023judging}
Lianmin Zheng, Wei-Lin Chiang, Ying Sheng, Siyuan Zhuang, Zhanghao Wu, Yonghao Zhuang, Zi Lin, Zhuohan Li, Dacheng Li, Eric Xing, et~al.
\newblock Judging llm-as-a-judge with mt-bench and chatbot arena.
\newblock \emph{Advances in Neural Information Processing Systems}, 36:\penalty0 46595--46623, 2023.

\bibitem[Zhou et~al.(2024)Zhou, He, Sun, Xu, Chen, Chu, Zhou, Liao, Zhang, Afvari, et~al.]{zhou2024pre}
Juexiao Zhou, Xiaonan He, Liyuan Sun, Jiannan Xu, Xiuying Chen, Yuetan Chu, Longxi Zhou, Xingyu Liao, Bin Zhang, Shawn Afvari, et~al.
\newblock Pre-trained multimodal large language model enhances dermatological diagnosis using skingpt-4.
\newblock \emph{Nature Communications}, 15\penalty0 (1):\penalty0 5649, 2024.

\end{thebibliography}
}
% \input{appen.tex}
\clearpage
\setcounter{page}{1}
\maketitlesupplementary

\section{SOAP Note Structure}
\label{sec:soap_structure}

\begin{figure}[htbp]
    \centering
    \includegraphics[width=0.5\textwidth]{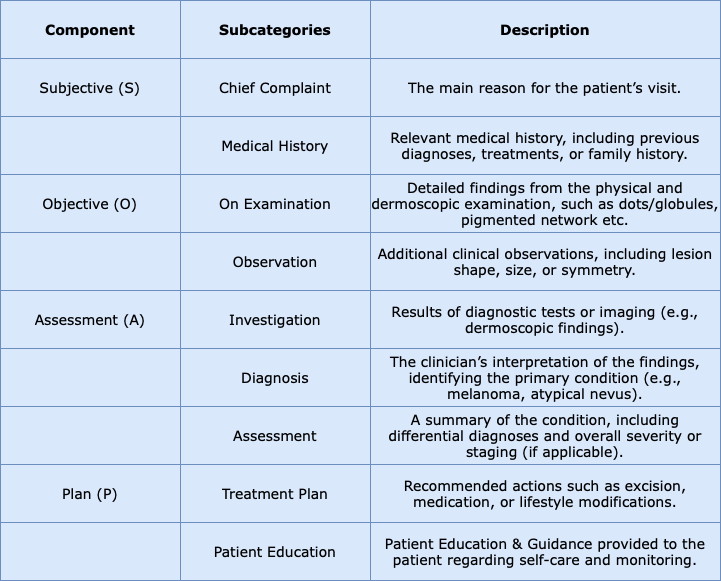}
    \caption{Structured SOAP note components.}
    \label{fig:SOAP_table}
\end{figure}

\section{Related Works}
\label{sec:related}

Artificial intelligence has been widely applied in the medical domain for tasks such as medical image captioning \cite{cheng2025caparena}, clinical text summarization \cite{van2023clinical}, discharge note generation \cite{jung2024enhancing}, and medical question answering \cite{yan2024large}. Recent efforts have focused on generating clinical notes, especially SOAP notes, from Electronic Health Records (EHRs) and doctor-patient conversations. In \cite{li2024improving}, large language models (LLMs) are fine-tuned to generate K-SOAP notes, while \cite{chen2024exploring} evaluated generative models across general and SOAP-specific formats. Similar work has explored domains like pediatric rehabilitation \cite{amenyo2025assessment} and fine-tuning LLMs for efficient SOAP note generation \cite{leong2024efficient}.

Earlier approaches to SOAP note generation, such as \cite{schloss2020towards,singh2023large}, used sequence-to-sequence models and transformer architectures trained on large dialogue datasets. Although general-purpose LLMs have improved clinical NLP tasks \cite{singhal2023large,yang2023large}, they often lack domain-specific reasoning and structured multimodal output capabilities.  While methods like \cite{ramprasad2023generating} have attempted to improve faithfulness in text-based settings, these methods heavily rely on large annotated dialogue corpora, which are difficult to obtain due to privacy concerns \cite{chen2024exploring}, and they remain text-centric without leveraging multimodal information, which is critical to fields like dermatology. SkinGPT-4 \cite{zhou2024pre} recently demonstrated that integrating clinical images and text using a multimodal LLM improves dermatological diagnostic reasoning. However, it is primarily designed for diagnostic prediction and lacks structured documentation capabilities. Specifically, it does not generate clinically formatted SOAP notes or support section-level reasoning required for clinical documentation. In contrast, our work focuses on end-to-end generation of structured SOAP notes from limited multimodal inputs, introducing weak supervision and domain-guided retrieval to ensure both clinical reliability and scalability.

% SkinGPT-4 \cite{zhou2024pre} recently demonstrated that integrating clinical images and text using a multimodal LLM improves dermatological diagnostic reasoning, but it does not produce structured documentation such as SOAP notes. Therefore, we propose a multimodal framework that takes only an input lesion image and related caption to generate structured SOAP notes, reducing reliance on large datasets including doctor-patient conversations and enhancing clinical reliability through domain-guided knowledge retrieval, particularly for skin lesion datasets.
\section{Methodology}
\label{sec:appendix method}
\subsection{Dataset}
\label{subsec:data}

We use the PAD-UFES-20 dataset~\cite{pacheco2020pad}, which consists of 2,298 dermoscopic images along with structured metadata for 1,641 skin lesions collected from 1,373 patients. The lesions are classified into six types: \textbf{Basal Cell Carcinoma (BCC)}, \textbf{Melanoma (MEL)}, \textbf{Squamous Cell Carcinoma (SCC, including Bowen’s disease)}, and three non-cancerous conditions: \textbf{Actinic Keratosis (ACK)}, \textbf{Seborrheic Keratosis (SEK)}, and \textbf{Nevus (NEV)}.

Approximately 58\% of the samples are confirmed through biopsy, while the remaining cases are clinically diagnosed based on expert consensus. Each lesion is linked to a CSV file containing 26 structured clinical features, which include lesion characteristics (such as size and anatomical location), patient demographics, symptom information (such as itching, bleeding, or changes in appearance), and family medical history.

\subsection{Parameter-Efficient Fine-Tuning (PEFT)}
\label{subsubsec:peft}

To reduce computational costs without compromising model performance, we employ Parameter-Efficient Fine-Tuning (PEFT) strategies. Specifically, we use Quantized Low-Rank Adaptation (QLoRA) ~\cite{dettmers2023qlora}, which introduces trainable low-rank matrices into transformer layers. Instead of updating all model parameters, QLoRA injects low-rank decompositions into specific modules such as the query, key, value, and output projections.

%The low-rank adaptation update is defined as:

% \begin{equation}
% \Delta W = A B
% \end{equation}

% \noindent
% where \(A \in \mathbb{R}^{d \times r}\) and \(B \in \mathbb{R}^{r \times d}\) are the learnable low-rank matrices and \(r \ll d\).
% The adapted weight matrix during fine-tuning becomes:

% \begin{equation}
% W' = W + \Delta W = W + A B
% \end{equation}

% \noindent
% where \(W\) represents the original frozen weights, and \(\Delta W\) captures the trainable adaptation.
% We apply LoRA to the \texttt{q}, \texttt{k}, \texttt{v}, \texttt{o}, \texttt{gate}, \texttt{up}, and \texttt{down} modules of the Vision-LLaMA model, allowing efficient adaptation without requiring full parameter updates.

% \begin{equation}
% \Delta W = AB
% \label{eq:deltaW}
% \end{equation}

% \noindent where \(A \in \mathbb{R}^{d \times r}\) and \(B \in \mathbb{R}^{r \times d}\) are the learnable low-rank matrices with \(r \ll d\).

% The adapted weight matrix during fine-tuning becomes:

% \begin{equation}
% W' = W + \Delta W = W + AB
% \label{eq:adaptedW}
% \end{equation}

% \noindent
% where \(W\) represents the original frozen weights, and \(\Delta W\) captures the trainable adaptation.
% We apply LoRA to the \texttt{q}, \texttt{k}, \texttt{v}, \texttt{o}, \texttt{gate}, \texttt{up}, and \texttt{down} modules of the Vision-LLaMA model, allowing efficient adaptation without requiring full parameter updates.

\subsection{Supervised Fine-Tuning (SFT)}
\label{subsubsec:sft}

Supervised Fine-Tuning (SFT) further adapts the model by explicitly teaching it to map dermatological inputs to structured SOAP notes based on weakly supervised ground-truth examples. Unlike pre-training, which loosely guides the model, SFT provides complete input-output pairs, enabling the model to learn structured clinical reasoning patterns.

The SFT training objective minimizes the cross-entropy loss between the predicted SOAP note \(\hat{y}\) and the target SOAP note \(y\) given the multimodal input \(x\) (image and caption):

\begin{equation}
\mathcal{L}_{\text{SFT}} = - \sum_{i=1}^{n} y_i \log(\hat{y}_i)
\end{equation}

\noindent
where \(\mathcal{L}_{\text{SFT}}\) denotes the supervised fine-tuning loss, \(n\) is the number of tokens in the SOAP note, \(y_i\) is the ground-truth token at position \(i\), \(\hat{y}_i\) is the predicted probability for the \(i\)-th token, and \(x\) represents the multimodal input consisting of the lesion image and its corresponding clinical caption. This loss function ensures that the generated notes are not only linguistically coherent but also structurally accurate and clinically reliable.

\subsection{Training Setup}
\label{subsec:training_setup}

The fine-tuning of the Vision-LLaMA 3.2 model is performed using Quantized Low-Rank Adaptation (QLoRA) with a low-rank dimension \( r = 8 \), a scaling factor \( \alpha = 16 \), and no dropout applied. LoRA modules are inserted into the query, key, value, output, gate, up, and down projections within the model's transformer blocks. Supervised Fine-Tuning (SFT) is conducted with a batch size of 8, using gradient accumulation over 4 steps and 10 warmup steps at the beginning of training. Fine-tuning is performed for 500 epochs with a linear learning rate scheduler, starting from an initial learning rate of \( 2 \times 10^{-4} \). The optimizer used is AdamW with 8-bit precision to improve memory efficiency. Pretraining to generate weakly supervised SOAP notes took approximately 9 hours, while the final fine-tuning process was completed in 1.5 hours on an NVIDIA A100 GPU with 80 GB of VRAM. Mixed-precision training with bfloat16 (bf16) format was employed to optimize memory utilization throughout the training process.

\section{Evaluation}
\label{evaluation}
\subsection{Quantitative Evaluation}
\label{sec:appendix quat}
\subsubsection{ROUGE}
ROUGE (Recall-Oriented Understudy for Gisting Evaluation) measures n-gram overlap between generated and reference notes. We report ROUGE-1 (unigrams), ROUGE-2 (bigrams), and ROUGE-L (longest common subsequence), capturing both lexical recall and structural consistency.

% \subsection{BLEU}
% BLEU (Bilingual Evaluation Understudy) evaluates n-gram precision, penalizing short or under-informative outputs. Due to the high variability in phrasing in clinical documentation, BLEU scores tend to be low, but still able to provide insight into exact overlap.

\subsubsection{METEOR}
METEOR (Metric for Evaluation of Translation with Explicit ORdering) incorporates synonymy and word stemming to better handle clinical paraphrasing. It is more tolerant to surface-level variations and is especially useful when evaluating partially reworded notes.

\subsubsection{CHRF++}
CHRF++ computes F-scores over character n-grams, offering robustness to minor lexical differences, such as pluralization, typos, or morphological variations (e.g., “carcinoma” vs. “carcinomas”). It is particularly well-suited for clinical domains where such variations are common.

\subsubsection{BERT Score}
BERT Score uses contextual embeddings from a pretrained language model (typically BERT-base) to compute semantic similarity between reference and generated notes. It evaluates whether generated tokens align semantically with reference tokens beyond exact matches.

\subsubsection{Clinical BERT Score}
To capture clinically grounded semantics, we use BERT with Clinical BERT, a model pretrained on large-scale clinical notes (e.g., MIMIC-III). This allows for more accurate evaluation of clinical relevance, capturing medical synonyms, abbreviations, and contextual language common in SOAP documentation.

\subsection{Qualitative Evaluation}
\label{sec:appendix qual}

\begin{figure*}[t]
    \centering
    \includegraphics[width=\textwidth]{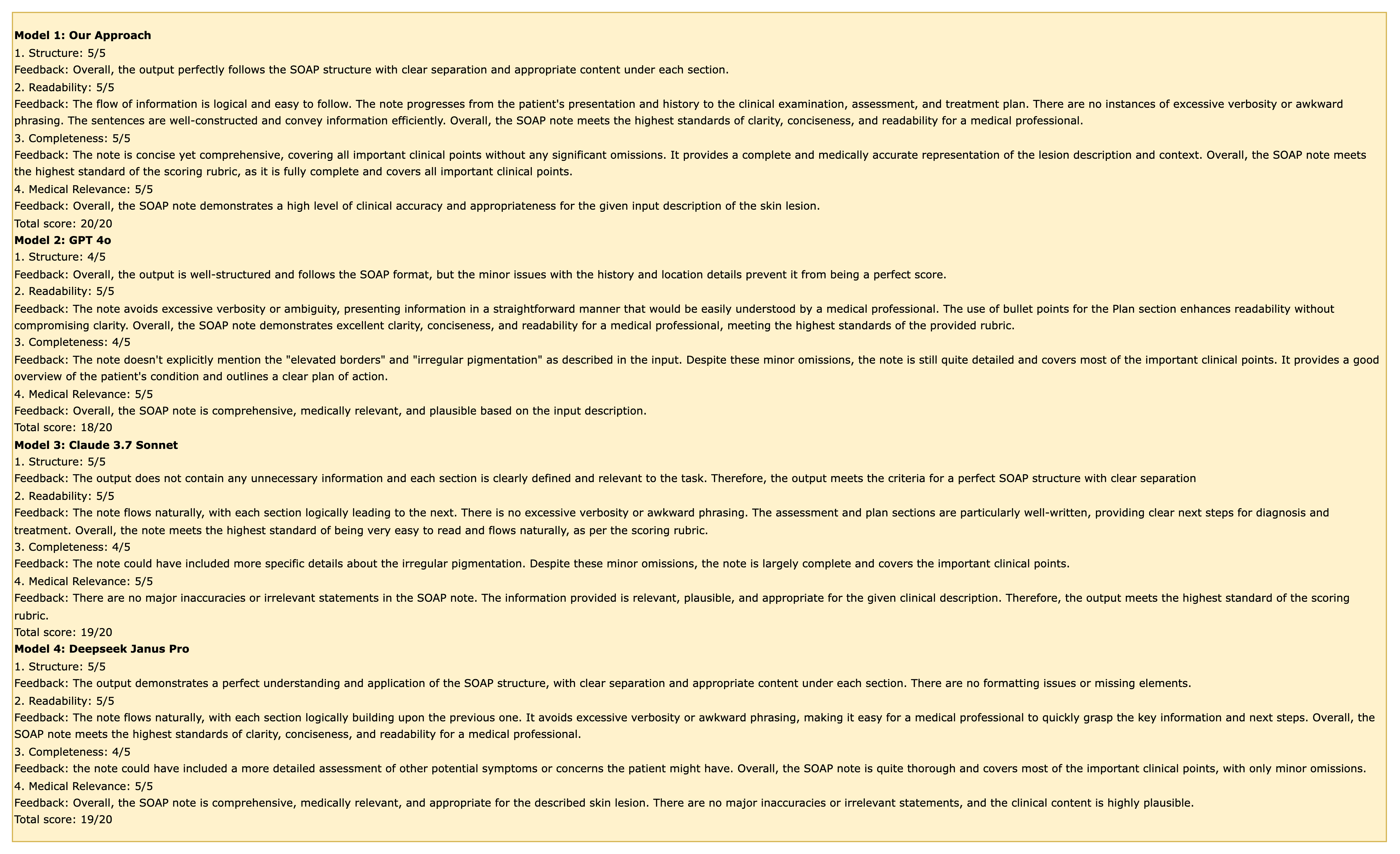}
    \caption{Comparison of Flow-Judge Feedback Across Four Language Models}
    \label{fig:LLM}
\end{figure*}

The generated notes were evaluated using Flow-Judge shown in \ref{fig:LLM} on a (1 = Poor, 5 = Excellent) rating scale based on the following criteria:

\begin{itemize} \item \textbf{Structure:} Does the clinical note correctly follow the structured SOAP format, with distinct and appropriate content under each section (Subjective: Chief Complaint and Medical History; Objective: Examination findings and Observations; Assessment: Investigations, Diagnosis, and Summary; Plan: Treatment Plan and Patient Education)? \item \textbf{Readability:} Is the language of the clinical note clear, concise, and readable for a medical professional without excessive complexity or ambiguity? \item \textbf{Completeness:} Does the clinical note cover all the key details described in the input lesion description and context, and address all aspects of the clinical scenario, ensuring that no critical details are overlooked? \item \textbf{Medical Relevance:} Is the clinical content of the SOAP note medically relevant, plausible, and appropriate given the input description of the skin lesion? \end{itemize}
% \section{Flow Judge Results}
% \label{sec:Flow judge}
% The total scores for each model were shown in Fig \ref{fig:LLM} ~\ref{sec:appendix qual} of the Appendix, our approach achieved a perfect score of 20/20, with feedback highlighting its clear organization, clinical accuracy, and comprehensive coverage of all relevant details. In contrast, GPT-4o and the other baselines scored between 18/20 and 19/20. While their outputs were well-structured and readable, minor issues such as omissions of specific symptoms or less detailed assessments slightly impacted their completeness scores. Overall, Flow-Judge evaluations confirm that our framework delivers higher clinical fidelity and completeness than SOTA models, highlighting key strengths and improvement areas.
\section{Example of Generated SOAP Notes}

\begin{figure*}[t]
    \centering
    \includegraphics[width=\textwidth]{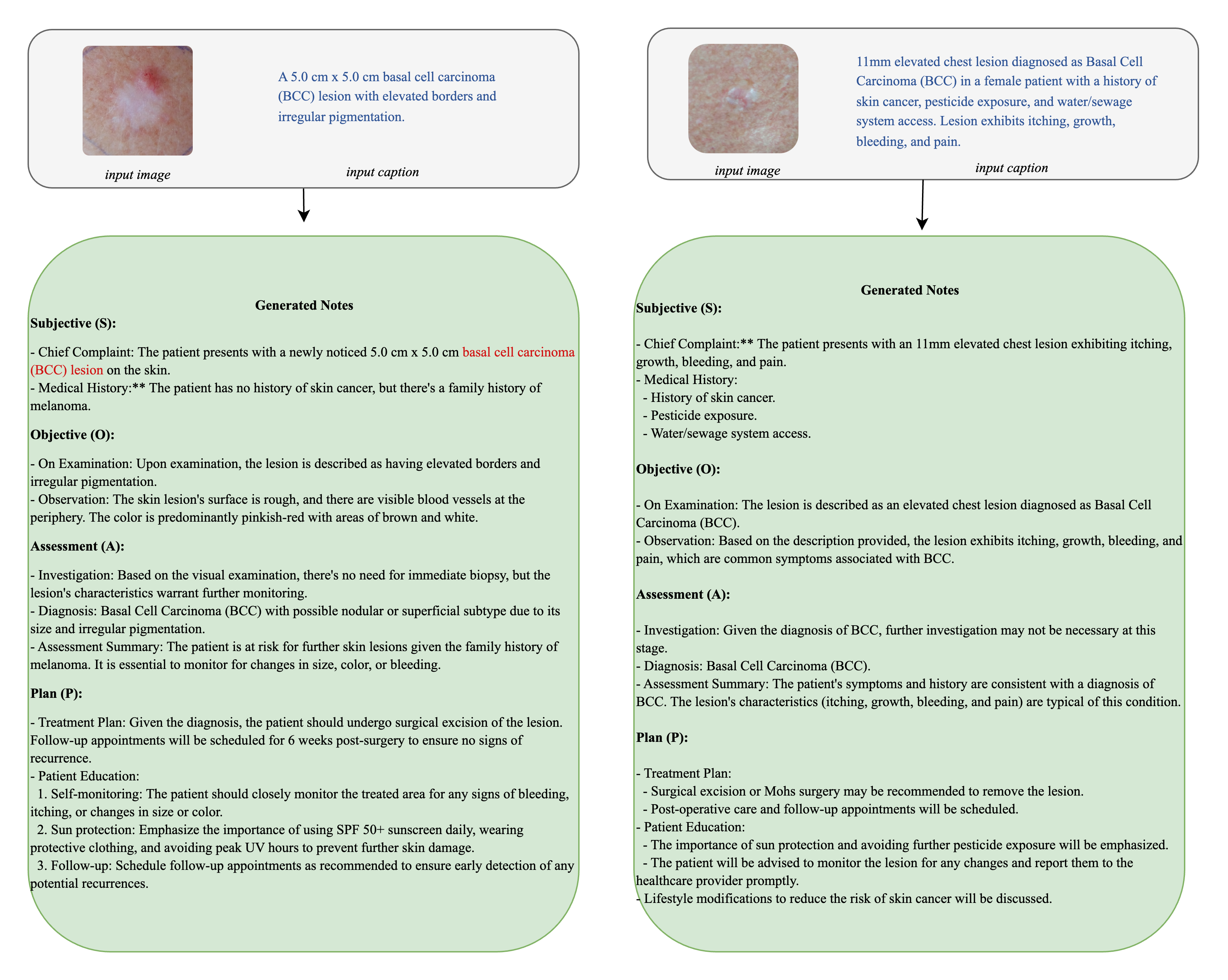}
    \caption{Examples (a) and (b) show structured SOAP notes generated by our proposed multimodal framework using an input lesion image and its corresponding caption.}
    \label{fig:result}
\end{figure*}

Figure~\ref{fig:result} presents two representative examples of structured SOAP notes generated by our proposed method, using an input lesion image and its corresponding caption. These examples highlights how the framework is able to produce clinically structured documentation aligned with the SOAP (Subjective, Objective, Assessment, Plan) format. As shown in the example, both cases, the model integrates information from the image and caption to synthesize comprehensive notes that reflect typical clinical reasoning and terminology. However, there are few cases where a model exhibits placement errors. As seen in Figure~\ref{fig:result}(a), the diagnosis Basal Cell Carcinoma (BCC) is prematurely introduced in the Chief Complaint section, which traditionally captures only the patient's presenting symptoms or concerns. This inclusion of diagnostic information at the subjective stage represents a structural inaccuracy, as it should be in the Assessment section.

Conversely, Figure~\ref{fig:result}(b) demonstrates a correctly formatted SOAP note. The Chief Complaint accurately reflects only the patient's reported symptoms upon arrival (e.g., itching, bleeding, pain), while the diagnosis of BCC is appropriately documented under Assessment, alongside supporting justification and summary. This example indicates that the model is capable of learning subtle distinctions in clinical documentation structure. While minor inconsistencies exist, these results affirm the potential of our approach to automate the generation of structured and clinically coherent SOAP notes from multimodal dermatologic inputs.

\section{Limitations and Future Work}
\label{sec:limit}

% While our weakly supervised multimodal framework shows strong promise for structured SOAP notes generation, it has few limitations. The quality of the generated SOAP notes remains dependent on the accuracy of the retrieved domain-specific knowledge, which may introduce biases or propagate incomplete information. Additionally, we utilized only a single dataset, as variations in metadata across different sources posed challenges for standardization, thereby limiting the model’s generalizability to broader clinical settings. Our evaluation was further constrained by a small set of expert-annotated samples, restricting large-scale validation. In Future work, we will focus on expanding to more diverse datasets, and incorporating human-in-the-loop refinement strategies. Furthermore, developing better evaluation benchmarks that track patient reasoning over time and support clinical decision-making could make automated clinical note generation much more useful in real-world healthcare settings.

While our weakly supervised multimodal framework shows strong promise for structured SOAP notes generation, it has few limitations. The quality of the generated SOAP notes remains dependent on the accuracy of the retrieved domain-specific knowledge, which may introduce biases or propagate incomplete information. Additionally, we utilized only a single dataset, as variations in metadata across different sources posed challenges for standardization. Our evaluation was further constrained by a small set of expert-annotated samples, restricting large-scale validation. Like most generative models, our approach may hallucinate when faced with ambiguous or unfamiliar inputs. Although retrieval-augmented generation helps reduce this risk by grounding outputs in clinical knowledge, additional safeguards are needed for real-world deployment. In Future work, we will focus on expanding to more diverse datasets, and incorporating human-in-the-loop refinement strategies. Furthermore, developing evaluation benchmarks that capture the progression of clinical reasoning across multiple encounters and support decision-making could significantly improve the utility of automated SOAP note generation in real-world healthcare settings.

% \section{Rationale}
% \label{sec:rationale}
% %
% Having the supplementary compiled together with the main paper means that:
% %
% \begin{itemize}
% \item The supplementary can back-reference sections of the main paper, for example, we can refer to \cref{sec:intro};
% \item The main paper can forward reference sub-sections within the supplementary explicitly (e.g. referring to a particular experiment);
% \item When submitted to arXiv, the supplementary will already included at the end of the paper.
% \end{itemize}
% %
% To split the supplementary pages from the main paper, you can use \href{https://support.apple.com/en-ca/guide/preview/prvw11793/mac#:~:text=Delete%20a%20page%20from%20a,or%20choose%20Edit%20%3E%20Delete).}{Preview (on macOS)}, \href{https://www.adobe.com/acrobat/how-to/delete-pages-from-pdf.html#:~:text=Choose%20%E2%80%9CTools%E2%80%9D%20%3E%20%E2%80%9COrganize,or%20pages%20from%20the%20file.}{Adobe Acrobat} (on all OSs), as well as \href{https://superuser.com/questions/517986/is-it-possible-to-delete-some-pages-of-a-pdf-document}{command line tools}.

\end{document}